\documentclass[11pt,a4paper]{article}
\usepackage[hyperref]{eacl2021}
\usepackage{times}
\usepackage{latexsym}
\usepackage{graphicx}
\usepackage{multirow}
\usepackage{tcolorbox}

\usepackage{microtype}

\aclfinalcopy 

\title{A preliminary study on evaluating Consultation Notes with Post-Editing\newline}

\author{
    Francesco Moramarco\thanks{~ Equal contribution} \\
  Babylon Health / London, UK \\
  University of Aberdeen / Aberdeen, UK \\
  \texttt{francesco.moramarco\textsuperscript{†}} \\\And
  Alex Papadopoulos Korfiatis\textsuperscript{*} \\
  Babylon Health / London, UK \\
  \texttt{alex.papadopoulos\textsuperscript{†}} \\
  \AND
  Aleksandar Savkov \\
  Babylon Health / London, UK \\
  \texttt{sasho.savkov\textsuperscript{†}} \\\And
  Ehud Reiter \\
  University of Aberdeen / Aberdeen, UK \\
  \texttt{e.reiter@abdn.ac.uk} \\
  \AND
  \textsuperscript{†}\normalfont{\texttt{@babylonhealth.com}}
}

\date{}

\begin{document}
\maketitle

\begin{abstract}

Automatic summarisation has the potential to aid physicians in streamlining clerical tasks such as note taking. But it is notoriously difficult to evaluate these systems and demonstrate that they are safe to be used in a clinical setting.
To circumvent this issue, we propose a semi-automatic approach whereby physicians post-edit generated notes before submitting them. 
We conduct a preliminary study on the time saving of automatically generated consultation notes with post-editing. Our evaluators are asked to listen to mock consultations and to post-edit three generated notes. We time this and find that it is faster than writing the note from scratch. We present insights and lessons learnt from this experiment.

\end{abstract}

\section{Introduction}
\label{sec:intro}

In modern EHR (Electronic Health Records) systems, at the end of a medical consultation the physician is required to file a consultation note detailing symptoms, examination, and treatment discussed. This is a pain point for physicians, who, according to a US study in 2017-2018 \cite{arndt2017tethered} spend up to $44.2\%$ of their time on clerical tasks, and this is a major contributor to physician burnout \cite{burnout}.

A number of recent studies \cite{kazi2019automatically, molenaar2020medical, krishna2020generating} propose to use summarisation systems to automatically generate the consultation note from the transcript of the consultation.
Yet there is limited work on how to properly evaluate such a system so that it may be used in the clinical setting. 
Intrinsic evaluation metrics through Likert scales or ranking methods may help select the best model, but they don't ensure the model will never hallucinate information or miss key items when generating the consultation note.

In this study we propose to evaluate generated consultation notes with an extrinsic measure based on post-editing time. We ask our evaluators (primary healthcare physicians) to listen to a consultation, write a consultation note, and post-edit a number of generated notes. We then compare the timings of each task to determine whether post-editing a note is faster than writing one from scratch. 

We focus on post-editing time because (i) it's simple to measure, and (ii) it provides a gate for adoption of the technology (i.e. post-editing a note should be faster than writing it from scratch). There are other extrinsic metrics which we intend to investigate in the future, such as patient satisfaction, doctor cognitive load, doctor-patient engagement, and usefulness for the next doctor accessing the note.

\section{Related Work}
\label{sec:related}

Post-editing has a long history in Machine Translation (MT) \cite{chander1994, carl2015post, graham2017improving, koponen2016machine, de2011assessing}, with a number of production systems and tools using a semi-automatic approach to fix errors and check the output of the system before it is shown to the users \cite{DowlingCltw16, aziz2012pet}.

Outside of MT, \citet{sripada2005evaluation} carry out a study on post-editing an NLG system for generating weather forecast from data.

As an evaluation metric, \citet{allman-etal-2012-linguists} define \emph{Productivity} as ``the quantity of text an experienced
translator could translate in a given period of time
[compared] with the quantity of text generated by
[the system] that the same person could edit in the
given time.''

To the best of our knowledge, post-editing is not widely used in document summarisation. We speculate this is partly because a post-editor of document summaries would need to read the entire document in order to accurately post-edit the summary, and this may minimise the benefit of having a generated summary compared to writing it from scratch.
This is not the case, however, with consultation note generation, whereby the physician in charge of writing the note is the same physician who has conducted the consultation. Here post-editing may be very valuable in saving physician time.



\section{Data}
\label{sec:data}

We partner with a UK healthcare provider, Babylon Health, which gives us access to a dataset of 800 proprietary consultation transcripts (automatically transcribed) and notes. The consultations span various topics within primary healthcare and are 10 minutes long on average. The notes are written by the physician who carried out the consultation and are in patient-friendly format, meaning they are in the same language the doctor used while talking to the patient and don't contain any abbreviations or acronyms. Each note is made up of three sections: \emph{History \& Examination}, \emph{Diagnosis}, and \emph{Management}.

For our evaluation, we design a dataset of 57 mock consultations produced in a similar manner.
We ask five Babylon Health physicians working in primary healthcare to act as doctors and a number of lay people (employees at Babylon) to act as patients.
Participation is entirely voluntary and all participants sign a consent form explaining what the study would involve and the intended use of the data produced. They are given the choice to withdraw consent at any point.

We give each patient a case card, prepared by a physician, that contains the condition they need help with and a list of medical details and symptoms.
We record the audio of each mock consultation and ask the doctor to write a patient-friendly note as described above. Figure \ref{fig:note-sample} shows a mock patient-friendly note.

\begin{figure}[t]

\begin{tcolorbox}[boxrule=1pt,arc=.2em,boxsep=0mm]
\textbf{HISTORY \& EXAMINATION}

You developed lower abdominal pain 2 days ago. The pain came on gradually, is burning in nature, constant and is worsening. You have no bowel symptoms or pain on urination, but have noticed a pink colour to your urine. You have not noticed and blood in your urine. You feel some nausea, but have not vomited. You feel hot and sweaty. You are sexually active with a long term partner. Your last sexual health check-up was 6 months ago. You last had unprotected sex 2 days ago. Your last period was 2 weeks ago. You have no other symptoms. You have no past medical history, but use implanon for contraception.

\vspace{3pt}

\textbf{DIAGNOSIS:} Urinary Tract Infection. Must rule out pregnancy

\vspace{3pt}

\textbf{MANAGEMENT}

Take a pregnancy test. Give urine sample for a urine dip and to check for bacteria. Treat with antibiotics. Regular paracetamol for pain. Review in 1 - 2 days if no improvement, or earlier if symptoms are worsening.

\end{tcolorbox}

    \caption{Mock consultation note written by a locum doctor, from our evaluation dataset.}
    \label{fig:note-sample}
\end{figure}

We then employ a transcription agency to transcribe the recordings on an utterance level. Figure \ref{fig:transcript-sample} shows a transcript snippet from the same consultation.

\begin{figure}[t]

\begin{tcolorbox}[boxrule=1pt,arc=.2em,boxsep=0mm]
[...]

\textbf{Doctor}: Hello? Good morning, Tim. Um, how can I help you this morning?

\textbf{Patient}: Um, so I'm having some, some pain, uh, in my tummy, like the lower part of my tummy. Um and I've just been feeling, quite, hot and sweaty.

\textbf{Doctor}: OK. Right, I'm sorry to hear that. When, when did your symptoms all start?

\textbf{Patient}: About two days ago.

\textbf{Doctor}: OK. And whereabouts in your tummy is the pain, exactly?

\textbf{Patient}: Uh, like below my belly button, it's like quite, sore when I press on it.

\textbf{Doctor}: OK. Did the pain come on quite suddenly, or was it more gradual?

\textbf{Patient}: it hasn't been, it's more gradual and it's just, it is getting a bit worse now.

\textbf{Doctor}: OK, OK. And can you describe the pain to me? [...]




\end{tcolorbox}

    \caption{Sample transcript from the mock dataset.}
    \label{fig:transcript-sample}
\end{figure}

For the evaluation reported here, we only use 3 out of the 57 consultations; we are planning to publish the whole dataset at a later stage.

\section{Experimental Setup}
\label{sec:experimental_setup}

We use the proprietary dataset of 800 consultations to finetune two automatic summarisation models based on BART \cite{lewis-etal-2020-bart}. We feed the transcripts as inputs and the consultation notes as outputs.

We then apply the models on the mock consultation dataset, using them to generate the \emph{History \& Examination} section of the consultation note. For our experiment, we consider the generated notes from these two models (Model A, Model B) together with the original reference note (Ref). We shuffle these for each task and tell the evaluators that all three notes are generated.

The task is presented to the evaluators using Heartex \cite{Heartex}, a configurable annotation platform that allows us to customise the design of the evaluation task.

Our evaluators are three primary healthcare physicians. They are employed at Babylon Health and have experience in AI research annotation. The task we submit to them consists of the following steps:
\begin{enumerate}
    \itemsep0em 
    \vspace{-0.05in}

    \item \textbf{Listen to the audio of a mock consultation.} We let evaluators note down any key symptoms on a piece of paper as they would normally do during a consultation.
    
    \item \textbf{Write the \emph{History \& Examination} sections of the consultation note (this is timed).} Just as they would in the clinical setting, after having listened to a consultation recording we ask them to write the first section of the consultation note. Figure \ref{fig:write-note} shows an example.

    \begin{figure}[t]
        \centering
        \includegraphics[width=.5\textwidth]{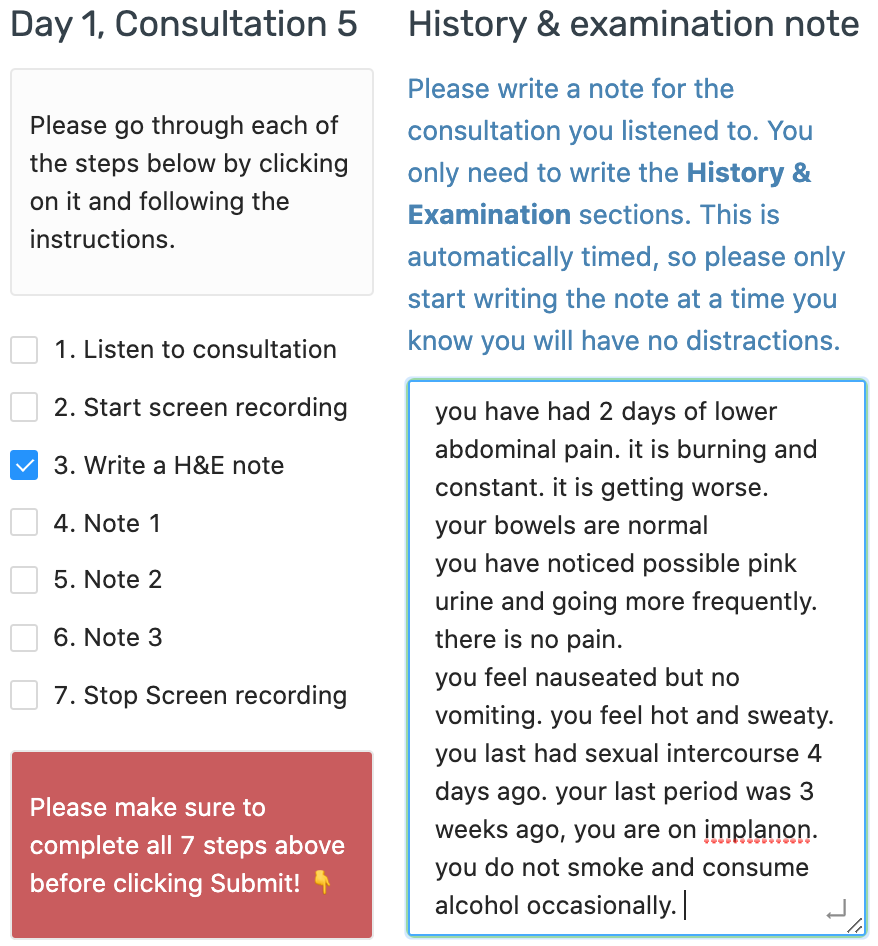}
        \caption{Heartex Annotation interface for writing the History\&Examination section of the consultation note.}
        \label{fig:write-note}
    \end{figure}
    
    \item \textbf{Post-edit three generated notes (this is timed).} The evaluators are presented with the three generated notes (Model A, Model B, Ref, in random order) for the given consultation and are asked to edit incorrect statements and to add missing statements.

\begin{figure*}[tbh]

\begin{tcolorbox}[boxrule=1pt,arc=.2em,boxsep=0mm]

\section*{Scoring Guidance}
We are scoring the quality of the note based on:

\vspace{0.05in}
\textbf{Correctness}: you will be asked to identify the number of incorrect statements in the note.
\vspace{0.05in}

\textbf{Completeness}: you will be asked to identify the number of major and minor omissions from the note. If an omission is negligible, please do not include it in the omission count. Here’s a description of each omission type:
\begin{itemize}
    \vspace{-0.07in}
    \item \textbf{Major} = any edit that would be needed before the consultation notes are completed (if not corrected, it would render the note unsatisfactory from a medico-legal and quality perspective) e.g. features of chest pain
        \vspace{-0.07in}

    \item \textbf{Minor} = any edit that would be preferable before the notes are completed (satisfied from a medico-legal point of view but deficient from a quality point of view) e.g. alcohol, smoking hx
    \vspace{-0.07in}

    \item \textbf{Negligible} = any edit if missed would not pose any issues but if included would improve the quality of the notes (this is information that you may tend not to record but if you had more time, you might record if you remember) e.g. medication hx which is already recorded elsewhere in the record
    \vspace{-0.07in}

\end{itemize}

\textbf{Coherence}: you will be asked if the note makes sense, regardless of the content.

\end{tcolorbox}

    \caption{Scoring guidance drafted by the lead physician.}
    \label{fig:scoring-guidance}
\end{figure*}

    We then present a number of questions to evaluate the quality of the given note. Our criteria are \emph{Correctness}, \emph{Completeness} \cite{goldstein2017evaluation}, and \emph{Coherence}. We agreed these criteria with the lead physician, who drafted definitions and a scoring guidance for the evaluators (Figure \ref{fig:scoring-guidance}).
    Figure \ref{fig:score-note} shows the questions we ask for scoring these criteria and a sample annotation.
    
    \begin{figure}[t]
        \centering
        \includegraphics[width=0.5\textwidth]{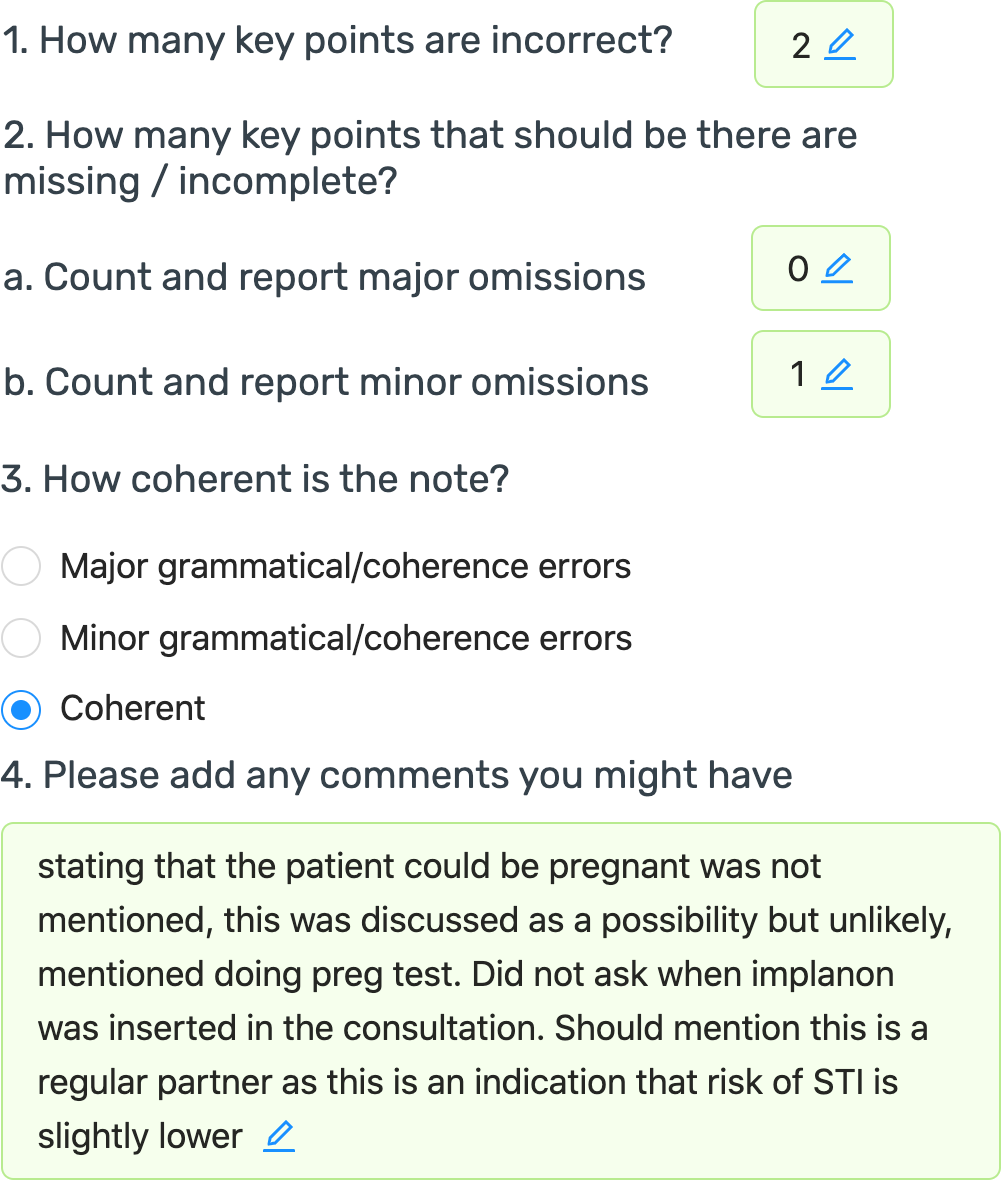}
        \caption{Heartex interface for scoring a generated note.}
        \label{fig:score-note}
    \end{figure}
    
    \vspace{-0.02in}

\end{enumerate}

We also ask evaluators to record their screen for the duration of the task. We use these recordings to calculate how long they took to write the note (step 2) and to edit each generated note (step 3). We use the difference of these two timings as our extrinsic measure to check whether editing a generated note is faster than writing one from scratch.


\begin{table*}[h]
\centering
\setlength{\tabcolsep}{4pt}
\begin{tabular}{l|lrrr|rrr|rrr|rrr}
\multicolumn{2}{l}{} & \multicolumn{3}{c}{\textbf{Incorrect}} & \multicolumn{3}{c}{\textbf{Major Omissions}} & \multicolumn{3}{c}{\textbf{Minor omissions}} & \multicolumn{3}{c}{\textbf{Coherence}} \\
\multicolumn{2}{l}{\textbf{Source}} & \multicolumn{1}{l}{\textbf{Dr A}} & \multicolumn{1}{l}{\textbf{Dr B}} & \multicolumn{1}{l}{\textbf{Dr C}} & \multicolumn{1}{l}{\textbf{Dr A}} & \multicolumn{1}{l}{\textbf{Dr B}} & \multicolumn{1}{l}{\textbf{Dr C}} & \multicolumn{1}{l}{\textbf{Dr A}} & \multicolumn{1}{l}{\textbf{Dr B}} & \multicolumn{1}{l}{\textbf{Dr C}} & \multicolumn{1}{l}{\textbf{Dr A}} & \multicolumn{1}{l}{\textbf{Dr B}} & \multicolumn{1}{l}{\textbf{Dr C}} \\ \hline
\multicolumn{2}{l}{\textbf{Ref}} & 0.67 & 2 & 1.67 & 0.33 & 0.67 & 1 & 0.33 & 3 & 0.67 & 2 & 2 & 1.67 \\
\multicolumn{2}{l}{\textbf{Model A}} & 1.67 & 2.33 & 1.33 & 0.67 & 3.67 & 3.33 & 1 & 4.67 & 0 & 2 & 1 & 1 \\
\multicolumn{2}{l}{\textbf{Model B}} & 1.67 & 2.33 & 0.67 & 1.67 & 3.33 & 3.33 & 0.33 & 5 & 1 & 1.67 & 1.67 & 1.33 \\ \hline
\end{tabular}
\caption{Aggregated scores for each evaluator, each criterion, and each note. For a full breakdown along tasks, please refer to Table \ref{tab:scores} in the Appendices.}
\label{tab:scores-aggregated}
\end{table*}

\section{Results and Discussion}
\label{sec:results}

For this experiment, we run our evaluation on $3$ of the $57$ mock consultations. Table \ref{tab:timings} gives a breakdown of the time it took to edit each note and write one from scratch. Here are some observations:

\begin{table}[t]
\centering
\setlength{\tabcolsep}{4pt}
\begin{tabular}{cc|c|ccc}
\textbf{Task} & \textbf{Eval} & \textbf{Write} & \textbf{Mod A} & \textbf{Mod B} & \textbf{Ref} \\ \hline
\multirow{4}{*}{\textbf{1}} & \textbf{Dr A} & 2:14 & 1:26 & 0:55 & 1:03 \\
 & \textbf{Dr B} & 4:02 & 4:30 & 4:16 & 2:44  \\
 & \textbf{Dr C} & 3:51 & 1:43 & 2:35 & 1:45\\ \cline{2-6} 
\multirow{4}{*}{\textbf{2}} & \textbf{Dr A} & 4:02 & 0:38 & 1:04 & 0:50\\
 & \textbf{Dr B} & 3:19 & 2:31 & 2:51 & 1:43 \\
 & \textbf{Dr C} & 2:26 & 1:10 & 1:16 & 0:42 \\ \cline{2-6} 
\multirow{4}{*}{\textbf{3}} & \textbf{Dr A} & 4:17 & 1:59 & 2:15 & 0:45 \\
 & \textbf{Dr B} & 4:21 & 4:04 & 3:32 & 4:17\\
 & \textbf{Dr C} & 3:53 & \multicolumn{1}{c}{-} & \multicolumn{1}{c}{-} & \multicolumn{1}{c}{-}\\
\end{tabular}
\caption{A breakdown of the time taken by the evaluators to write the note from scratch and post-edit each of the generated notes (Mod A, Mod B, Ref). The timings are in M:ss for minutes and seconds taken.}
\label{tab:timings}
\end{table}

\begin{itemize}
\itemsep0em 
    \vspace{-0.05in}

    \item In almost all cases, post-editing an existing note is faster than writing a note from scratch;
    
    \item As expected, post-editing the reference note (written by the consulting physician) is in general faster than post-editing the notes generated by either model. However, there are a number of instances (across all evaluators) where this isn't the case; 
    
    \item Note-taking style and length is very different amongst physicians \cite{cohen2019variation}, and this can be seen in our results as well. Doctor A tends to write shorter, terser notes and only edits the generated notes when there are substantial issues. Doctor B on the other hand is more meticulous and edits the generated notes extensively. This is reflected in both their edit times and their note scoring (see Table \ref{tab:scores-aggregated}). We report a detailed view of this disagreement in Figure \ref{fig:disagreement} in the Appendices;
    
    \item While it's not feasible to compute correlation between post-editing times and note scores given our sample size, there does seem to be a connection between the two: notes that are scored as containing more omissions and/or incorrect statements take longer to edit. For example, both Dr. B's aggregated scores (Table \ref{tab:scores-aggregated}) and edit timings (Table \ref{tab:timings}) are higher than the other two doctors.
    \item In one instance, one physician was so frustrated by the quality of a specific generated note that they decided to copy the note they wrote from scratch and paste it instead of trying to edit the generated one. This is why we have missing values in Table \ref{tab:timings};

    
    
    \item The first task each physician completed took 36 minutes on average, while subsequent tasks were quicker (23 minutes on average).
    
    \vspace{-0.02in}

\end{itemize}

After watching the recordings and collecting the results, we asked the three evaluators for qualitative feedback regarding the task, the annotation platform, and the generated notes. Here are the key insights we gathered:

\begin{itemize}
\itemsep0em 
    \vspace{-0.05in}

    \item Unlike post-editing, scoring is hard and time-consuming. This is partly due to the interface, which currently doesn't highlight the evaluators' edits on the generated note;

    \item Familiarity with the interface is key. We shadowed 2 of the 3 physicians through their first few tasks, and that reduced confusion and sped up their work. The physician we did not shadow expressed more difficulty in the evaluation task;

    \item Our evaluation setup — with physicians asked to listen to a consultation before writing the note — doesn't exactly reproduce the reality of the clinical setting, where they are actually conducting the consultation;
    
    \item One physician expressed the worry that even though post-editing a generated note might take less time than writing a note from scratch, it however requires a higher cognitive load. This is because the physician needs to critically read, understand and evaluate the generated note in order to correct it.
    
    \item In our experiment, we always ask the evaluators to first write a note from scratch, and then post-edit the generated notes. This specific order may bias our timings. The evaluators may be faster in post-editing after having written the note, or they may be slower if the generated note doesn't follow their style of writing. We plan to address this by shuffling the order of these two tasks. 
    
    
        \vspace{-0.02in}

\end{itemize}



\section{Future work}
\label{sec:future-work}

In this paper, we presented our preliminary evaluation study of consultation note generation with post-editing. Based on the insights from this study, we plan to: 

\begin{itemize}
\itemsep0em 
    \vspace{-0.05in}

    \item Extend the evaluation to the entire mock consultation dataset and calculate agreement between the evaluators. It would also be interesting to compute agreement between the scores (Correctness, Completeness) and the time taken to post-edit;

    \item Evaluate the usefulness of auto-generated notes in a live clinical setting;

    

    \item Investigate and compare the cognitive load of post-editing notes with that of writing them.
    
    \vspace{-0.02in}

\end{itemize}


If the issues described in this paper are addressed, we believe post-editing time can be a metric that is both valuable for evaluating model performance and relevant for use in production systems.

Finally, it is important to mention that while automation of medical note taking might help reduce physician burnout and allow the doctors to spend more time with the patients, there are ethical considerations associated to the use of such a technology.
For example, time pressures or unwarranted trust in an automated system could potentially result in doctors not properly reviewing and editing the automated notes. 
Also, post-editing is a very different cognitive task from writing a note from scratch, and that might put extra strain on doctors' already cognitively demanding workflows.
In order to mitigate the above concerns in a production system, user experience design, system evaluation, and clinician on-boarding and training are crucially important.

\bibliography{anthology,eacl2021}
\bibliographystyle{acl_natbib}

\newpage
\onecolumn
\appendix
\setcounter{table}{0}
\renewcommand{\thetable}{A\arabic{table}}
\setcounter{figure}{0}
\renewcommand{\thefigure}{A\arabic{figure}}

\section{Appendices}
\label{sec:appendix}

\begin{table*}[h]
\setlength{\tabcolsep}{4pt}
\begin{tabular}{llllllllllllll}
 &  & \multicolumn{3}{c}{\textbf{incorrect}} & \multicolumn{3}{c}{\textbf{major omissions}} & \multicolumn{3}{c}{\textbf{minor omissions}} & \multicolumn{3}{c}{\textbf{coherence}} \\
\multicolumn{2}{l}{\textbf{Task \& Source}} & \textbf{Dr A} & \textbf{Dr B} & \textbf{Dr C} & \textbf{Dr A} & \textbf{Dr B} & \textbf{Dr C} & \textbf{Dr A} & \textbf{Dr B} & \textbf{Dr C} & \textbf{Dr A} & \textbf{Dr B} & \textbf{Dr C} \\ \hline
\multirow{3}{*}{\textbf{1}} & \textbf{ref} & 2 & 2 & 1 & 1 & 0 & 1 & 0 & 4 & 1 & 2 & 2 & 2 \\
 & \textbf{model A} & 1 & 2 & 1 & 0 & 3 & 4 & 2 & 6 & 0 & 2 & 1 & 2 \\
 & \textbf{model B} & 0 & 1 & 1 & 2 & 4 & 4 & 0 & 6 & 1 & 2 & 1 & 1 \\ \hline
\multirow{3}{*}{\textbf{2}} & \textbf{ref} & 0 & 2 & 3 & 0 & 1 & 0 & 0 & 3 & 0 & 2 & 2 & 1 \\
 & \textbf{model A} & 1 & 2 & 1 & 0 & 3 & 2 & 0 & 5 & 0 & 2 & 1 & 1 \\
 & \textbf{model B} & 2 & 1 & 1 & 0 & 3 & 2 & 1 & 5 & 0 & 2 & 2 & 1 \\ \hline
\multirow{3}{*}{\textbf{3}} & \textbf{ref} & 0 & 2 & 1 & 0 & 1 & 2 & 1 & 2 & 1 & 2 & 2 & 2 \\
 & \textbf{model A} & 3 & 3 & 2 & 2 & 5 & 4 & 1 & 3 & 0 & 2 & 1 & 0 \\
 & \textbf{model B} & 3 & 5 & 0 & 3 & 3 & 4 & 0 & 4 & 2 & 1 & 2 & 2
\end{tabular}
\caption{Scores table.}
\label{tab:scores}

\end{table*}

\begin{figure*}[tbh]
    \centering
    \includegraphics[width=1.0\textwidth]{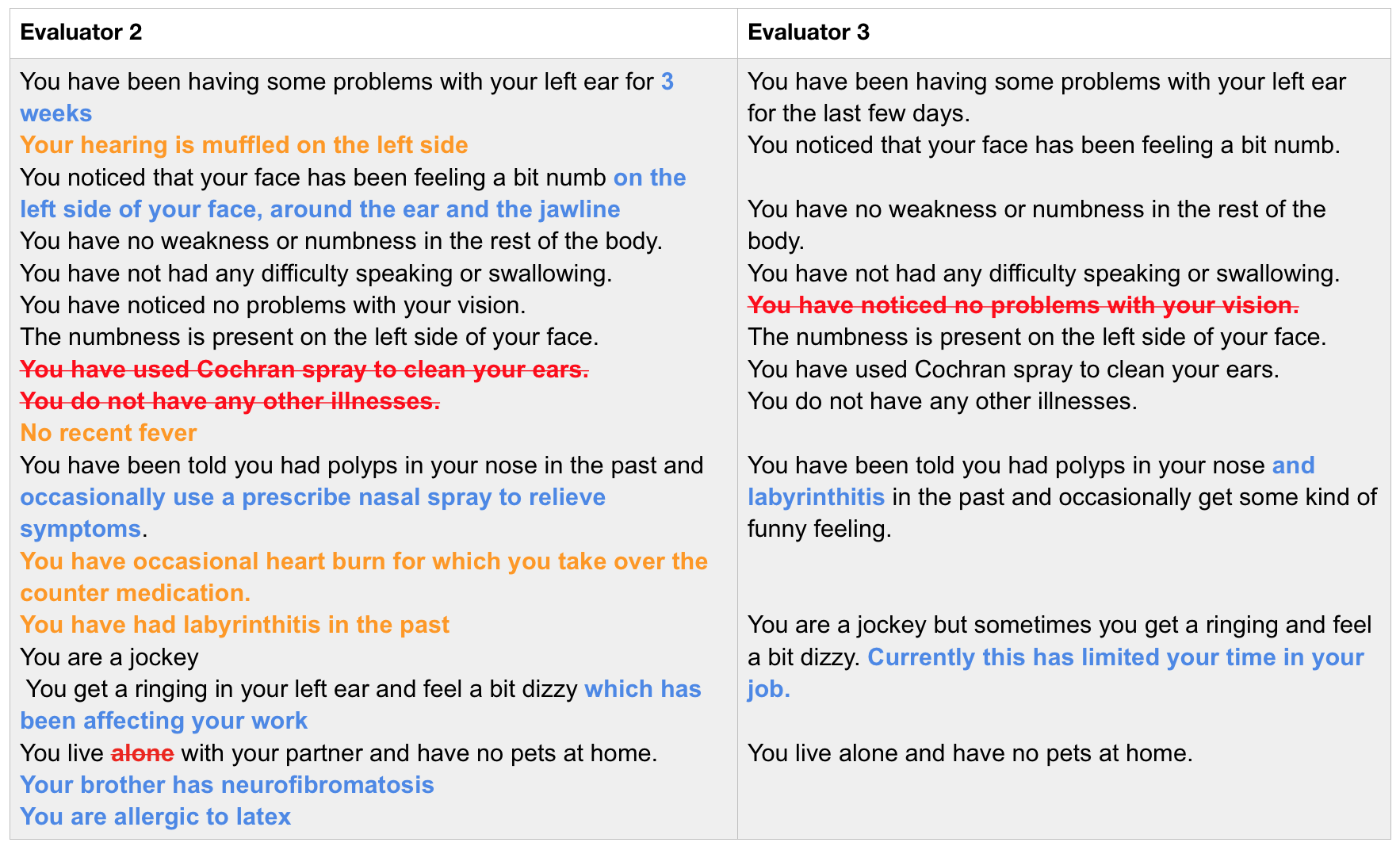}
    
    \caption{Disagreement in editing and scoring a generated note. \textcolor{red}{\textbf{Red}} marks incorrect statements, \textcolor{orange}{\textbf{orange}} major omissions, and \textcolor{blue}{\textbf{blue}} minor omissions.}
    \label{fig:disagreement}
\end{figure*}


\end{document}